\documentclass{article}
\usepackage{spconf,amsmath,graphicx}

\usepackage{graphicx} 
\usepackage{multirow} 
\usepackage[table]{xcolor} 
\usepackage{soul} 
\usepackage{hyperref}
\usepackage[numbers,sort&compress]{natbib}
\usepackage[utf8]{inputenc}
\usepackage{balance} 
\usepackage{array, boldline, makecell, booktabs} 

\title{PAG-Net: Progressive Attention Guided Depth Super-resolution Network}
\name{Author(s) Name(s)\thanks{Thanks to CSIR for funding.}}
\address{Author Affiliation(s)}

\twoauthors
  {Arpit Bansal$^{\ast}$, Sankaraganesh Jonna$^{\dagger}$, and Rajiv R. Sahay$^{\ast}$}
	{Department of Electrical Engineering$^{\ast}$, Department of Computer Science and Engineering$^{\dagger}$\\
	Indian Institute of Technology Kharagpur, West Bengal, India}
{}
{}

\begin{document}
%
\maketitle
\begin{abstract}
In this paper, we propose a novel method for the challenging problem of guided depth map super-resolution, called PAG-Net. It is based on residual dense networks and involves the attention mechanism to suppress the texture copying problem arises due to improper guidance by RGB images. The attention module mainly involves providing the spatial attention to guidance image based on the depth features. We evaluate the proposed trained models on test dataset and provide comparisons with the state-of-the-art depth super-resolution methods.
\end{abstract}
\section{Introduction}
\label{sec:intro}

A geometric description of a scene, high-quality depth map is useful in many computer vision applications, such as 3D reconstruction, virtual reality, scene understanding, intelligent driving, and robot navigation.
Literature mainly contains two classes of techniques for depth information acquisition, which are passive methods and active sensors. Firstly, passive methods infer depth maps from the most widely used dense stereo matching algorithms \cite{Szeliski_pami2008_stereo} but they are time-consuming. Despite the advances in technology, the depth information from passive methods are still inaccurate in occluded and low-texture regions. The acquisition of high-quality depth maps are more challenging to obtain than RGB images.

Depth acquisition from active sensors has become increasingly popular in our daily life and ubiquitous to many consumer applications, due to their simplicity, portability, and inexpensive. Unlike passive methods, depth of a scene can be acquired in real-time, and they are more robust in low-textured regions by low-cost sensors such as Time-of-Flight camera \cite{Lange_JQE2001_ToF} and Microsoft Kinect \cite{Herrera_pami2012_Kinect}. Active sensing techniques measures depth information of a scene by using echoed light rays from the scene. Time-of-Flight sensor (ToF) is one of the a mainstream type which computes depth at each pixel between camera and subject, by measuring the round trip time. Although depth-sensing technology has attracted much attention, it still suffers from several quality degradations.

Depth information captured by ToF sensors suffer from low-spatial resolutions (e.g., $176\times 144$, $200\times 200$ or $512\times 424$) and noise when compared with the corresponding color images. Due to the offset between projector and sensor, depth maps captured by Microsoft Kinect sensors contains structural missing along discontinuities and random missing at homogeneous regions. 
These issues restrict the use of depth maps in the development of depth-dependent applications. High-quality depth is significant in many computer vision applications. Therefore, there is a need for restoration of depth maps before using in applications. In this work, we consider the problem of depth map super-resolution from a given low-resolution depth map and its corresponding high-resolution color image. 

\begin{figure*}[!htb]
	\centering
	\begin{tabular}{c}
		\includegraphics[width=17cm]{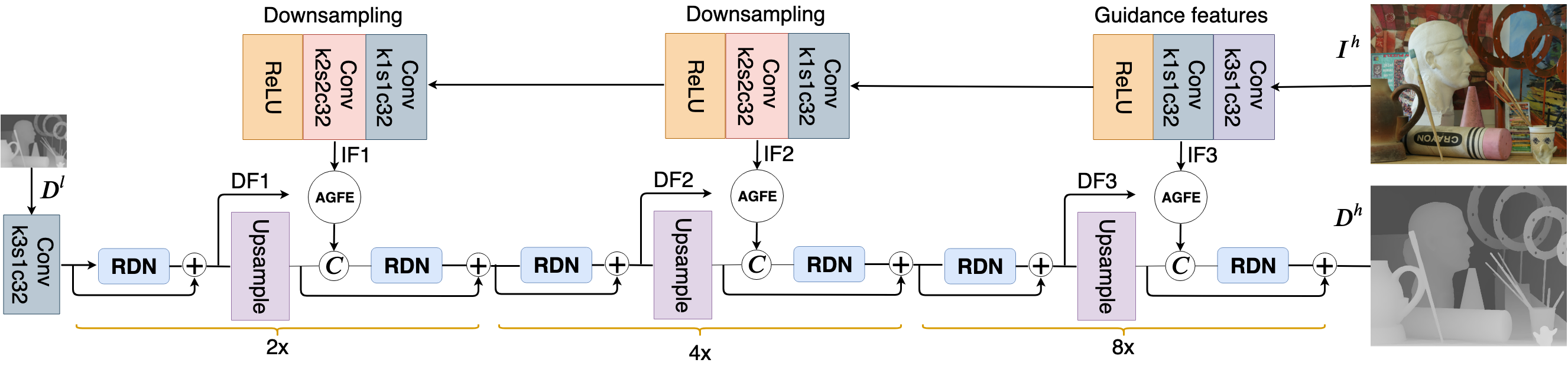}\\	
		(a)\\		
	\end{tabular}	
	\caption{Architecture of the proposed progressive attention guided depth super-resolution (PAG-Net) network for $8$x upsampling.}
	\label{fg:archi}
\end{figure*}

Existing depth super-resolution (DSR) methods can be roughly categorized into three groups: filter design-based \cite{He_pami2013}, optimization-based \cite{Ferstl_iccv2013,Yang_tip2014,Park_tip2014,Xie_tip2016,Liu_tip2017,Jiang_tip2018}, and learning-based algorithms \cite{hui_eccv16,Song_tcsvt2019,Wen_tip2019,Zuo_tcsvt2019,Zuo_IS2019,Guo_tip2019}. Many of the existing methods assumed that corresponding high-resolution color image helps improving the quality of depth maps and used aligned RGB image as guidance for depth SR.
However, major artifacts including texture copying and edge blurring, may occur when the assumption violated. 
Color texture would be transferred to the super-resolved depth maps if the smooth surface contains rich textures in the corresponding color image. Secondly, depth and color edges might not align in all the cases, subsequently, it leads to ambiguity. Hence, there is a need of optimal guidance for the high-resolution depth map.

Although there have been many algorithms \cite{Ferstl_iccv2013,Jiang_tip2018,hui_eccv16,Zuo_tcsvt2019,Zuo_IS2019,Guo_tip2019} proposed in the literature for the depth super-resolution (DSR), most of them still suffer from edge-blurring and texture copying artifacts.
In this paper, we propose a novel method for attention guided depth map super-resolution. It is based on residual dense networks \cite{He_cvpr2016,Zhang_cvpr2018} and involves novel attention mechanism. The attention used here to suppress the texture copying problem arises due to improper guidance by RGB images and transfer only the salient features from guidance stream. The attention module mainly involves providing the spatial attention to guidance image based on the depth features. The entire architecture for the example of super-resolution by the factor of $8$ is shown in Fig. \ref{fg:archi}. 

\section{Methodology}
\label{sec:method}

\begin{figure}[!htb]
	\centering
	\begin{tabular}{c c}
		\includegraphics[width=2.2cm]{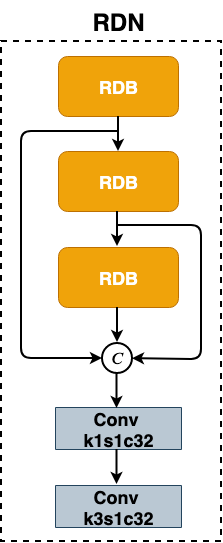}&\hspace{-12pt}	
		\includegraphics[width=4.5cm]{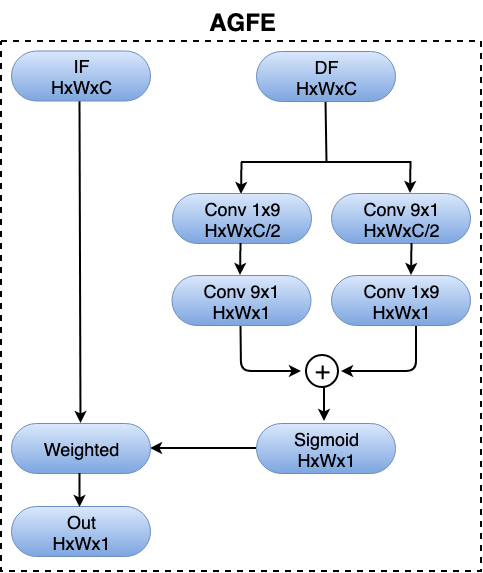}\\		
		(a) & (b)\\		
	\end{tabular}	
	\caption{(a) RDN: Residual Dense Network. (b) AGFE: Attention Guided Feature Extraction module.}
	\label{fg:archi_blocks}
\end{figure}

We design a deep progressive attention guided depth super-resolution network (PAG-Net) to learn to relationship between low-resolution depth map $D^{l}\in q\times H\times W$ and its corresponding high-resolution depth $D^{h} \in H\times W$ with the help of guidance image $I^{h} \in H\times W\times 3$. It consists of two streams, a progressive depth upsampling branch and multiscale guidance feature extraction branch shown in Fig. \ref{fg:archi}. Inspired by the image super-resolution work in \cite{Zhang_cvpr2018}, for the fusion of multi-level features, we consider residual dense network (RDN) with three residual dense blocks (RDB) as a building block in our depth super-resolution stream shown in Fig. \ref{fg:archi_blocks} (a). Specifically, the proposed PAG-Net model contains a novel attention guided feature (AGFE) extraction module shown in Fig. \ref{fg:archi_blocks} (b). The attention module mainly involves providing the spatial attention to guidance image based on the depth features. The entire architecture for the example of super-resolution by the factor of $8$ shown in Fig. \ref{fg:archi}. 


Most of the CNN architectures face the problem of gradient vanishing as the deeper networks can’t propagate the gradient deeper. This was handled by ResNets \cite{He_cvpr2016} by signal passing from one layer to another. There are other networks too which do the same. Moreover in most of them some information is lost as we propagate information from one layer to another. DenseNets \cite{Huang_cvpr2017} which actually ensure maximum flow of learned information and hence ensures that no information is lost and properly utilized. This also results in shallower architectures. Recently, residual dense networks (RDN) \cite{Zhang_cvpr2018} have been proved to be successful for the problem of image super resolution which ensures both suppressing gradient vanishing problem and helps in integrating multi-level features. In our depth upsampling network , we have treated RDN as a basic module it consists of $3$ residual dense blocks which are connected concurrently as shown in the Fig. \ref{fg:archi_blocks} (a). The output of each RDB is then concatenated and passed through the convolution layer of kernel size $1$ establishing a global dense connection. Finally, the features are passed through the convolution layer. The same architecture of RDN has been used in all the places. 

\textbf{Attention Guided Feature Extraction:}
The guidance of RGB images for the super-resolution of depth map is what sets the problem apart from color image super-resolution. Due to this guidance, depth super-resolution produces results even for the factor of $16$. The main problem faced due to improper guidance is of texture copying and smooth discontinuities. It becomes more prominent as the factor of upsampling increases. Hence, the guidance signal requires an attention mechanism to minimize the texture copying and allows to transfer only consistent salient edges. The overall architecture of attention module is in Fig. \ref{fg:archi_blocks} (b). 

%

\begin{figure}[!htb]
	\centering
	\begin{tabular}{c c c c}
		\includegraphics[width=2.0cm]{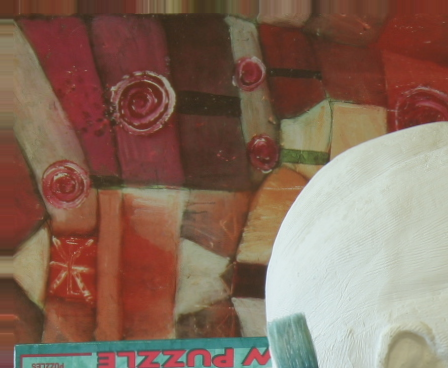}&\hspace{-12pt}	
		\includegraphics[width=2.0cm]{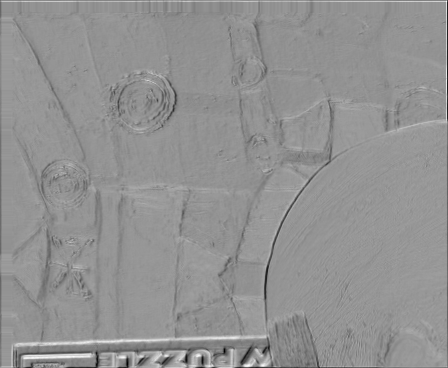}&\hspace{-12pt}	
		\includegraphics[width=2.0cm]{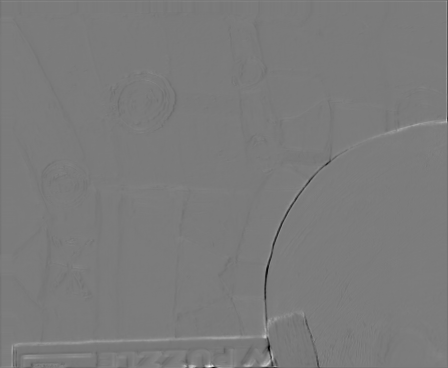}&\hspace{-12pt}
		\includegraphics[width=2.0cm]{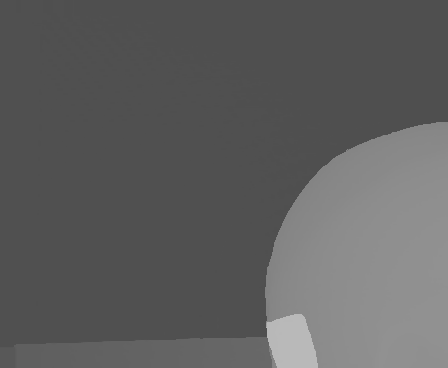}\\	
		(a) & (b) & (c) & (d)\\		
	\end{tabular}	
	\caption{(a) Guidance image. (b), (c) Feature map from guidance stream without and with proposed attention. (d) Ground truth depth map.}
	\label{fg:features}
\end{figure}

The guidance for depth super-resolution is its corresponding color image and being a natural image it has a lot of information in it. Some of its information such as edges present in them is useful for depth super-resolution. But not all the edges are important for example, if there is a painting on the wall, the edges in the painting are not useful, while the edges of the wall are crucial. This results in texture copying in depth super resolution. Hence there is need for selecting consistent and salient features from guidance channel. This is where the spatial attention steps in. It gives more weightage to the regions which are more probable to be copied in the depth image. In Fig. \ref{fg:features} (a), we show the guidance color image. The features obtained guidance stream without and with using proposed AGFE module are shown in Figs. \ref{fg:features} (b), (c), respectively. High-resolution depth profile corresponding to (a) is shown in Fig. \ref{fg:features} (a). Note that the proposed AGFE module transfers only the salient features required for the depth map super-resolution.


We represent the input depth features as $DF\in H\times W\times C$ and it is passed through $2$ different sets of convolution layers. First set involves a convolution layer of kernel size $9\times 1$ which is followed by a convolution layer of kernel size $1\times 9$. The second set is the reverse of the first stage, that is at first there is a convolution layer of kernel size $1\times 9$ and then a layer of kernel size $9\times 1$. This is done to increase the receptive field while at the same time not increasing the number of parameters. In the above two sets, at first the number of channels are reduced to $C/2$ and then decreased to $1$. Finally the output of the two sets are added and passed through the sigmoid layer which generates the spatial attention weights.


For the robust recovery of depth map at large upsampling factors the proposed PAG-Net upsamples the depth in multiple stages instead of direct super-resolution. At every stage, the depth features are upsampled by the factor of $2$ and its corresponding guidance features gets downsampled by $2$. AGF at every stage takes mutual information from both depth and intensity features and transfers only salient information for depth map upsampling by suppressing inconsistent artifacts.  
The guidance image passes through series of downsampling, in which at each stage it gets downsampled by a factor of $2$. Each downsampling stage receives the guidance features and at first it passes through the generic convolution layer and then it gets downsampled by a convolution operation with kernel size $2$ and stride $2$. For the example of $8$x , we get guidance features IF$3$ directly from $I^{h}$, IF$2$ downsampled by a factor of $2$, and IF$1$ downsampled by a factor of $4$. The first stage hence takes IF$3$ and depth features produced by passing depth image through a generic convolution layer followed by RDNs, AGFE and upsampling modules, and produces features upsampled by a factor of $2$.


In the proposed architecture, each stage can be further divided into $2$ sub-stages. The first sub-stage deals with upsampling the depth image features produced by the previous stage. The second sub-stage involves the fusion of guidance features attentively and followed by refinement. First stage takes low-resolution depth map as its input and before entering the RDN it passes through the generic convolution layer. RDN is then followed by global residual connection which produces the feature set DF. DF then passes through the upsampling layer which increases the spatial-resolution by $2$x. The upsampling module which we used here is a pixel shuffle layer and produces the feature set at high-resolution. The architecture for the second sub-stage, which involves the fusion of guidance features attentively with the upsampled depth features. Before passing through the attention module, the guidance features are passed through the generic convolution layer. The output of the attention module is then concatenated with upsampled features and then passed through the convolution layer of kernel size $3$. This in turn is passed through RDN and is followed by the global residual connection. This stage-wise upsampling process repeats for $l$ number of times for the upsampling factor $2^l$.

	\begin{table*}[!htb]
		\centering
		\caption{Qunatitative comparison on noise-free Middlebury dataset $2003$ \cite{Scharstein_ijcv2002} in RMSE with four upsampling factors.}
		\scalebox{0.64}
		{
			\begin{tabular}{|c|cccc|cccc|cccc|cccc|cccc|cccc|}
				\hline
				\multirow{2}{*}{Methods} & 
				\multicolumn{4}{|c|}{Art} & 
				\multicolumn{4}{|c|}{Books} &
				\multicolumn{4}{|c|}{Laundry} &
				\multicolumn{4}{|c|}{Reindeer} &
				\multicolumn{4}{|c|}{Tsukuba} &
				\multicolumn{4}{|c|}{Teddy} \\\cline{2-25}
				& 2x & 4x & 8x & 16x & 2x & 4x & 8x & 16x & 2x & 4x & 8x & 16x & 2x & 4x & 8x & 16x & 2x & 4x & 8x & 16x & 2x & 4x & 8x & 16x \\
				\hline
				Bicubic & 2.64 & 3.87 & 5.46 & 8.17 & 1.07 & 1.61  & 2.34 & 3.34 & 1.61 & 2.41 & 3.45 & 5.07 & 2.64 & 3.87 & 5.46 & 8.17 & 1.07 & 1.61  & 2.34 & 3.34 & 1.61 & 2.41 & 3.45 & 5.07\\
				GF \cite{He_pami2013} & 3.15 & 3.90 & 5.43 & 8.15 & 1.43 & 1.76 & 2.38 & 3.34 & 2.01 & 2.47 & 3.44 & 5.03 & 3.15 & 3.90 & 5.43 & 8.15 & 1.43 & 1.76 & 2.38 & 3.34 & 2.01 & 2.47 & 3.44 & 5.03\\
				TGV \cite{Ferstl_iccv2013} & 3.16 & 3.73 & 7.12 & 12.08 & 1.33 & 1.67 & 2.27 & 4.89 & 1.87 & 2.25 & 4.05 & 8.01 & 3.16 & 3.73 & 7.12 & 12.08 & 1.33 & 1.67 & 2.27 & 4.89 & 1.87 & 2.25 & 4.05 & 8.01\\
				JID \cite{Kiechle_iccv2013}  & 1.18 & 1.92 & 2.76 & 9.93 & 0.45 & 0.71 & 1.01 & 8.43 & 0.68 & 1.10 & 1.83 & 8.73 & 1.18 & 1.92 & 2.76 & 9.93 & 0.45 & 0.71 & 1.01 & 8.43 & 0.68 & 1.10 & 1.83 & 8.73\\
				MSG \cite{hui_eccv16}  & 0.66 & 1.47 & 2.46 & 4.57 & 0.37 & 0.67 & 1.03 & 1.63 & 0.79 & 0.79 & 1.51 & 2.63  & 0.66 & 1.47 & 2.46 & 4.57 & 0.37 & 0.67 & 1.03 & 1.63 & 0.79 & 0.79 & 1.51 & 2.63\\
				RCG \cite{Liu_tip2017}  & 2.31 & 3.26 & 4.31 & 6.78 & 1.14 & 1.53 & 2.18 & 2.92 & 1.47 & 2.06 & 2.87 & 4.22 & 1.82 & 2.58 & 3.24 & 4.90 & - & - & - & - &-  & - & - & -\\	
				TSDR \cite{Jiang_tip2018} & 3.16 & 1.57 & 2.30 & 4.30 & 1.33 & 1.05 & 1.06 & 1.59 & 1.87 & 0.98 & 1.58 & 2.19 & 3.16 & 1.57 & 2.30 & 4.30 & 1.33 & 1.05 & 1.06 & 1.59 & 1.87 & 0.98 & 1.58 & 2.19\\			
				MFR \cite{Zuo_tcsvt2019} & 0.71 & 1.54 & 2.71 & 4.35 & 0.42 & 0.63 & 1.05 & 1.78 & 0.61 & 1.11 & 1.75 & 3.01 & 0.65 & 1.23 & 2.06 & 3.74 & - & - & - & - & - & - & - & -\\
				RDN \cite{Zuo_IS2019} & 0.56 & 1.47 & 2.60 & 4.16 & 0.36 & 0.62 & 1.00 & 1.68 & 0.48 & 0.96 & 1.63 & 2.86 & 0.51 & 1.17 & 2.05 & 3.58 & - & - & - & - & - & - & - & -\\
				DSR \cite{Guo_tip2019} & 0.53 & 1.20 & 2.22 & 3.90 & 0.42 & 0.60 & 0.89 & 1.51 & 0.44 & 0.78 & 1.31 & 2.26 & 0.51 & 0.96 & 1.57 & \textbf{2.47} & 1.33 & 3.26 & 6.89 & 13.10 & 0.83 & \textbf{1.37} & \textbf{1.85} & 3.02\\					
				\hline
				PAG-Net & \textbf{0.33} & \textbf{1.15} & \textbf{2.08} & \textbf{3.68} & \textbf{0.26} & \textbf{0.46} & \textbf{0.81} & \textbf{1.38} & \textbf{0.30} & \textbf{0.71} & \textbf{1.27} & \textbf{1.88} & \textbf{0.31} & \textbf{0.85} & \textbf{1.46} & 2.52 & \textbf{0.85} & \textbf{3.08} & \textbf{6.18} & \textbf{11.84} & \textbf{0.66} & 1.39 & 1.90 & \textbf{2.91}\\
				\hline
			\end{tabular}			}
			\label{table:sr_quant}
		\end{table*}

	\begin{figure*}[!htb]
		\centering
		\scalebox{1.0}{
			\begin{tabular}{c c c c c c c}
				\includegraphics[width=2.4cm]{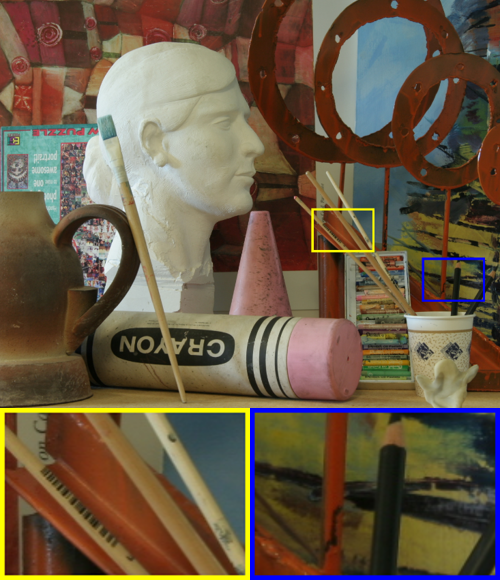}&\hspace{-13pt}	
				\includegraphics[width=2.4cm]{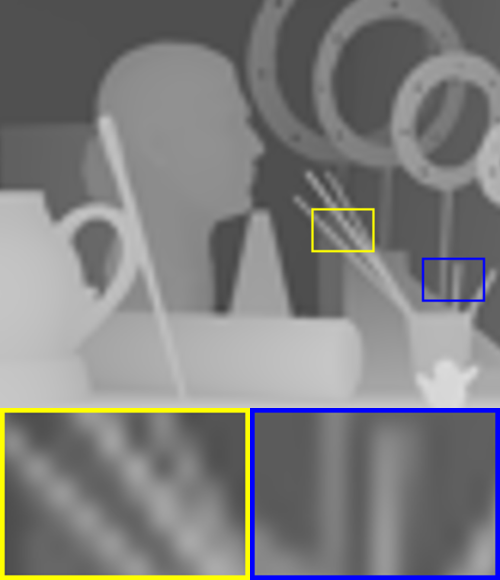}&\hspace{-13pt}
				\includegraphics[width=2.4cm]{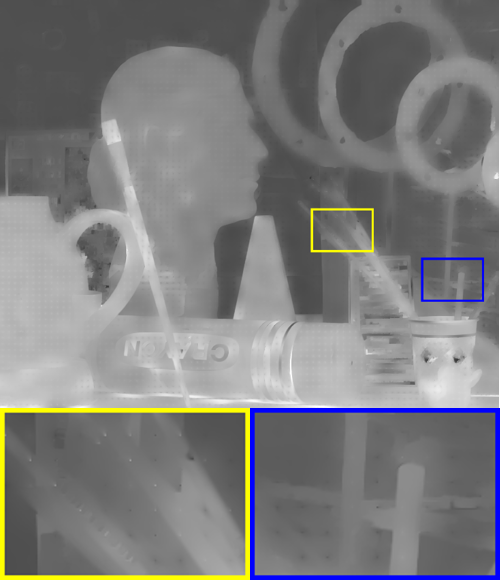}&\hspace{-13pt}	
				\includegraphics[width=2.4cm]{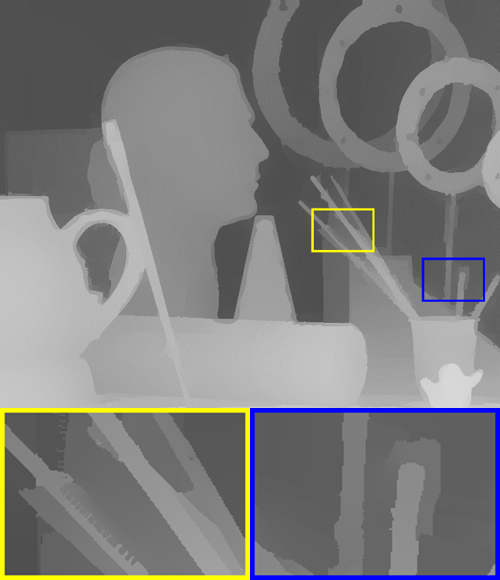}&\hspace{-13pt}							
			    \includegraphics[width=2.4cm]{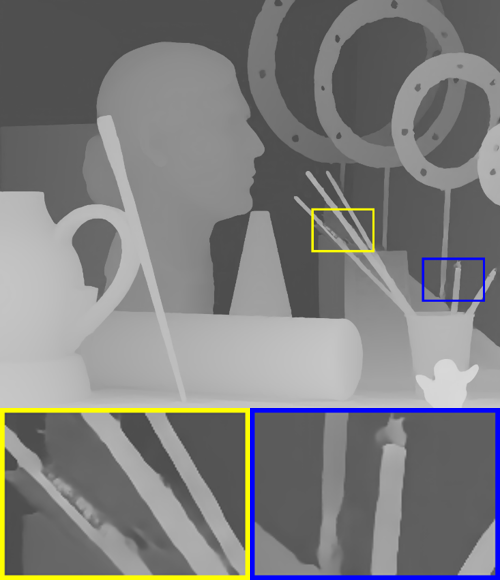}&\hspace{-13pt}		
				\includegraphics[width=2.4cm]{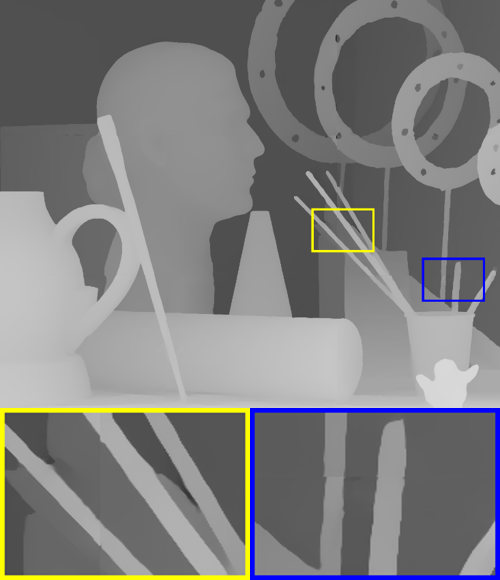}&\hspace{-13pt}
				\includegraphics[width=2.4cm]{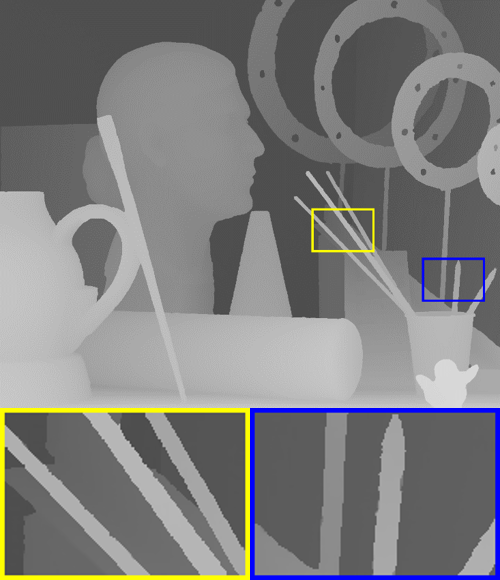}\\	
				\includegraphics[width=2.5cm]{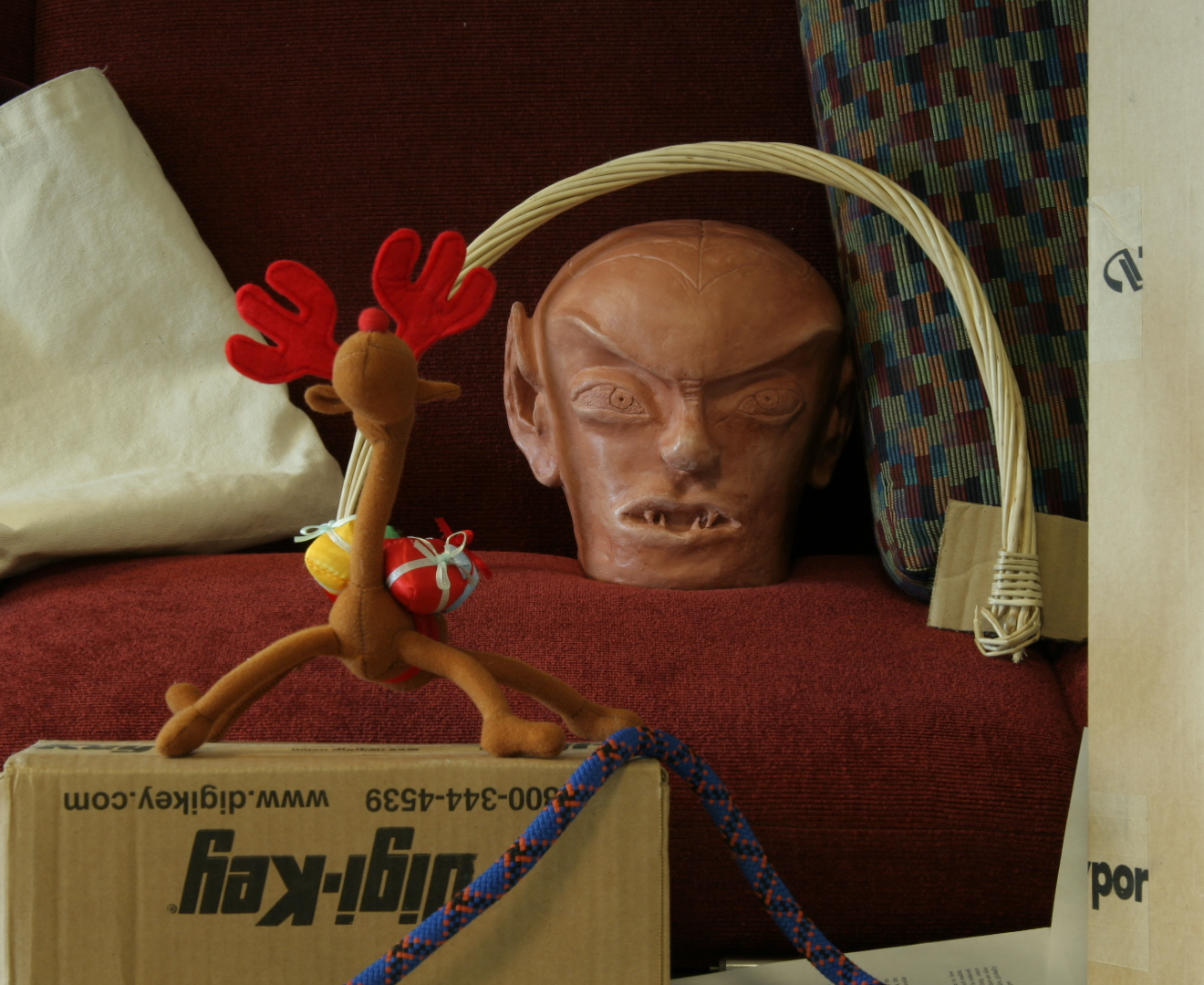}&\hspace{-13pt}
				\includegraphics[width=2.5cm]{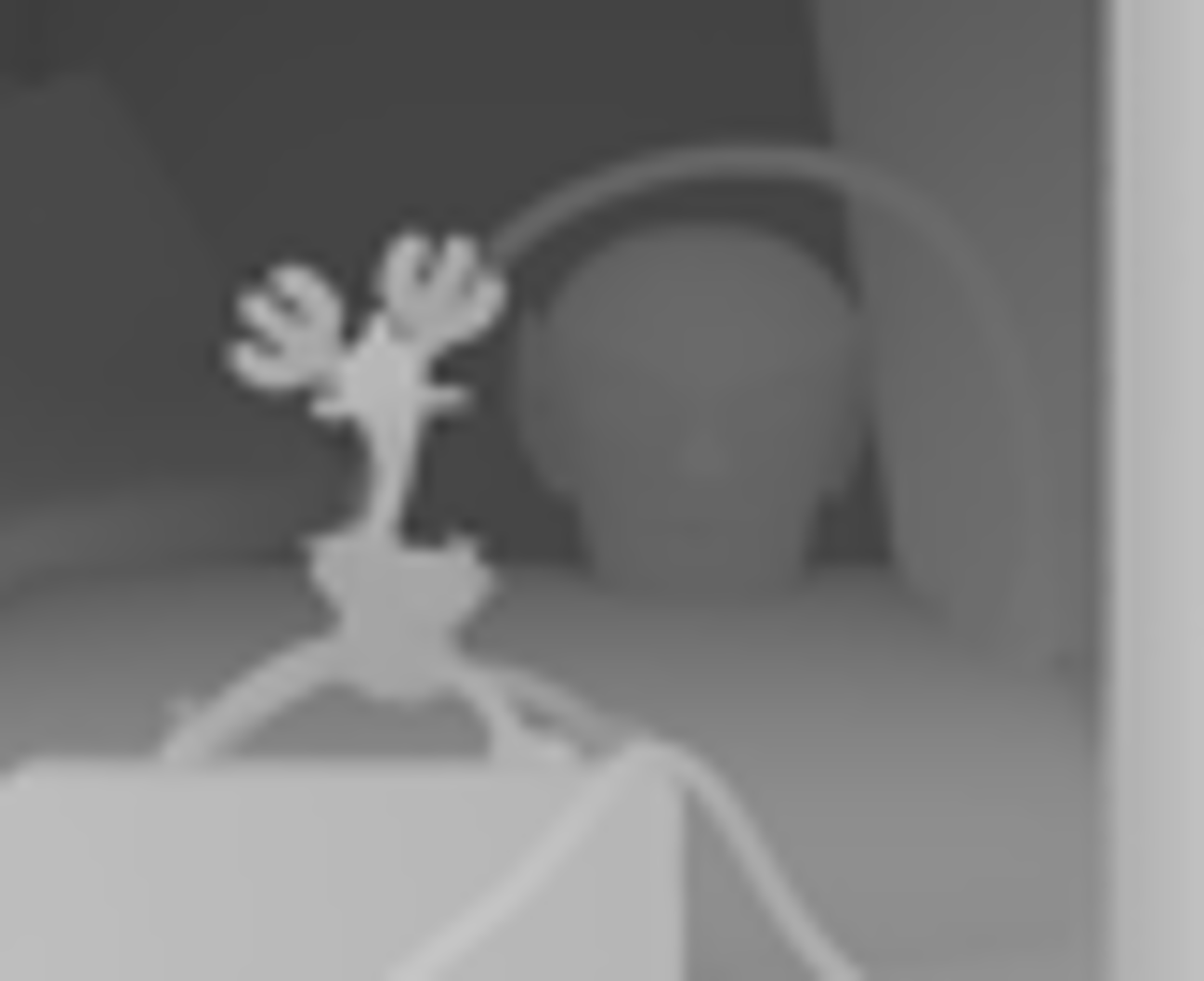}&\hspace{-13pt}
				\includegraphics[width=2.4cm]{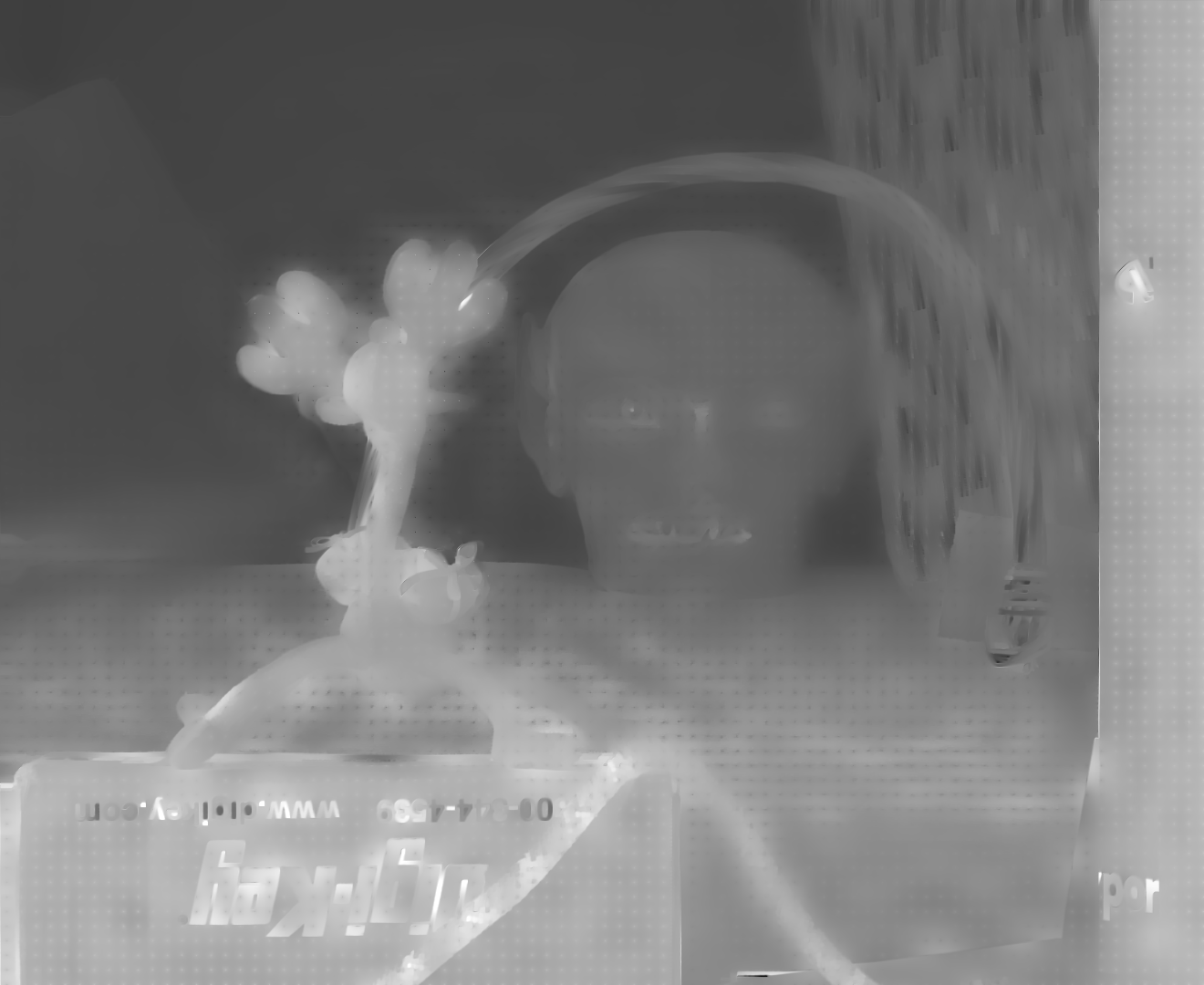}&\hspace{-13pt}	
				\includegraphics[width=2.4cm]{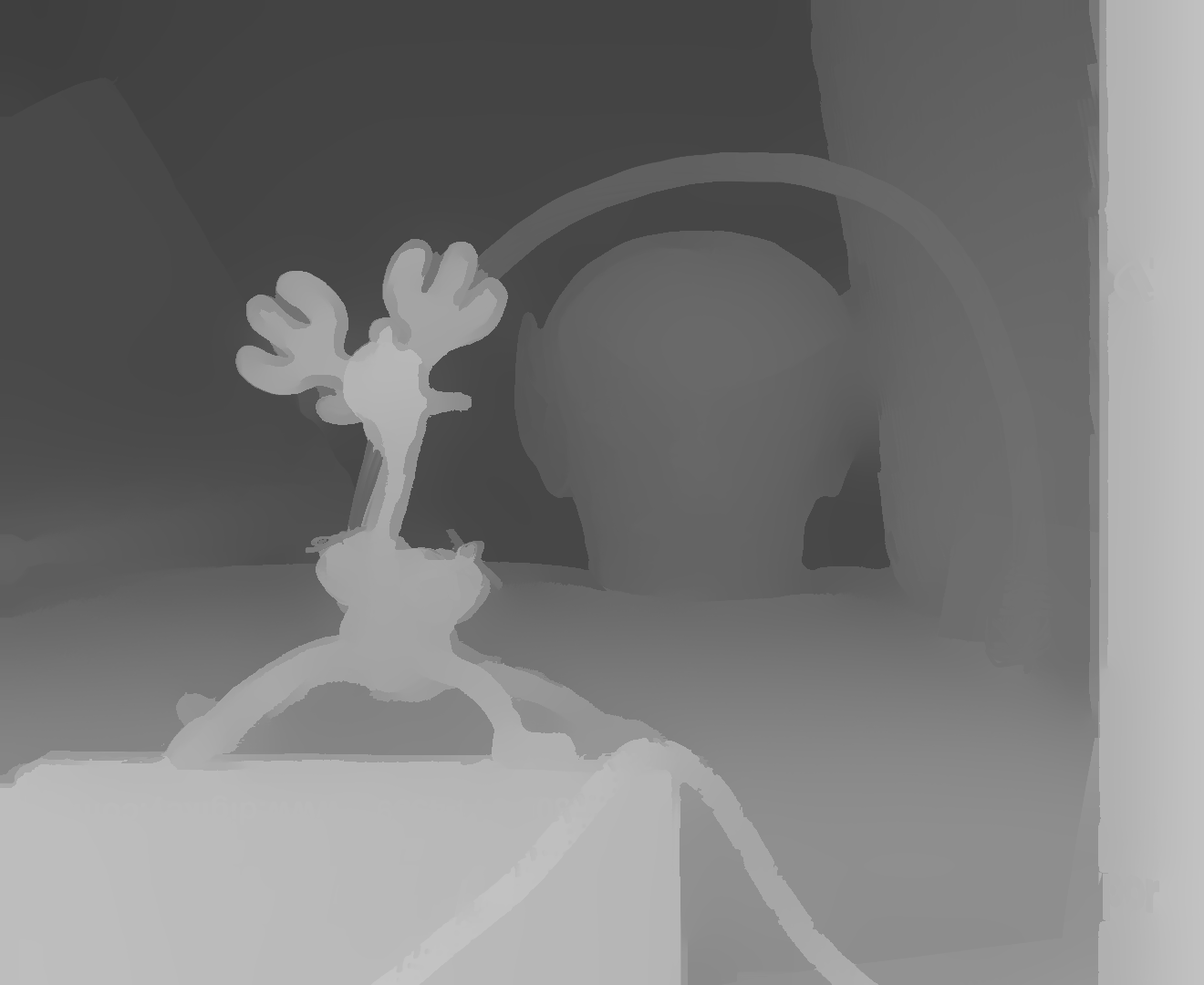}&\hspace{-13pt}				
				\includegraphics[width=2.5cm]{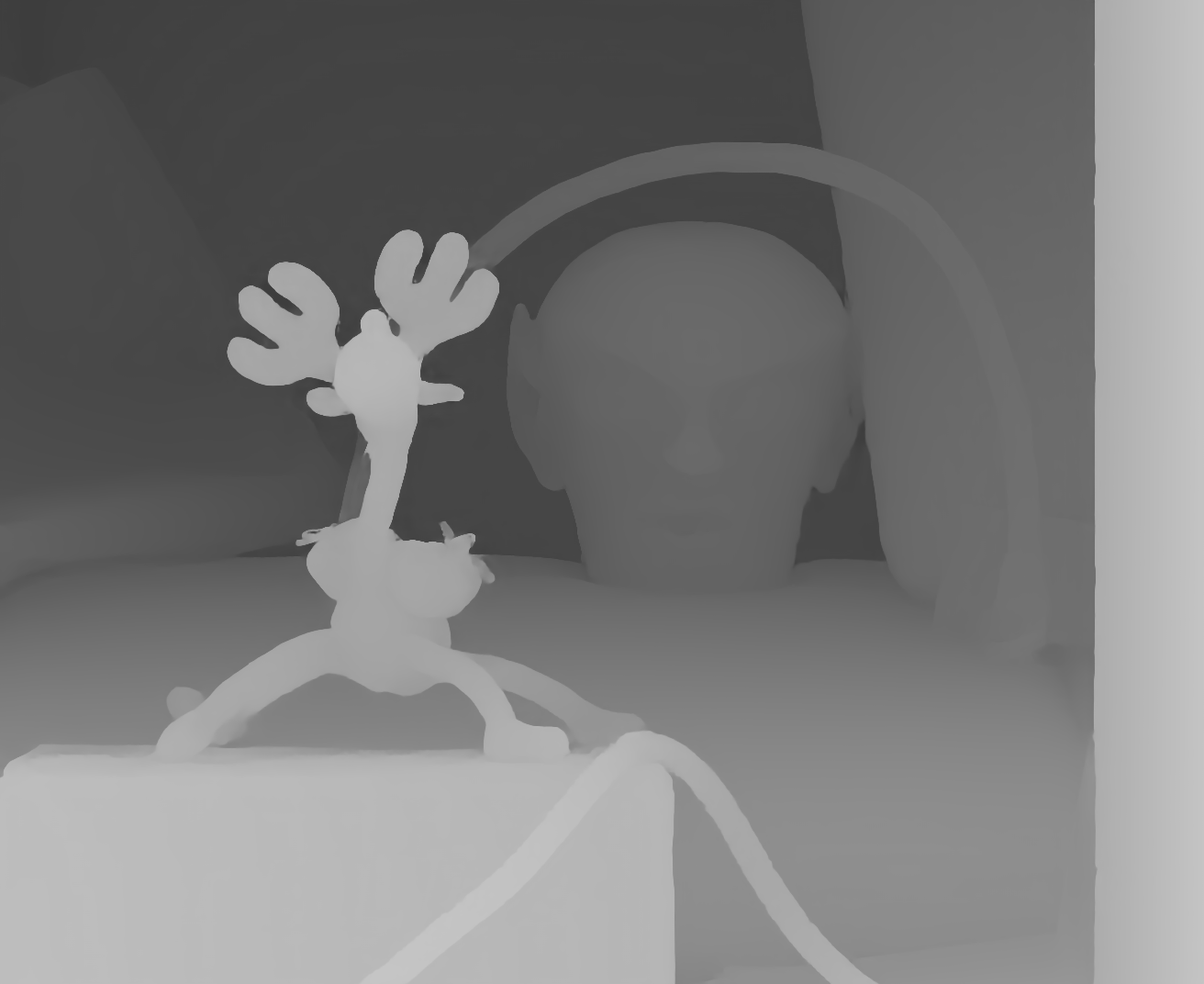}&\hspace{-13pt}		
				\includegraphics[width=2.5cm]{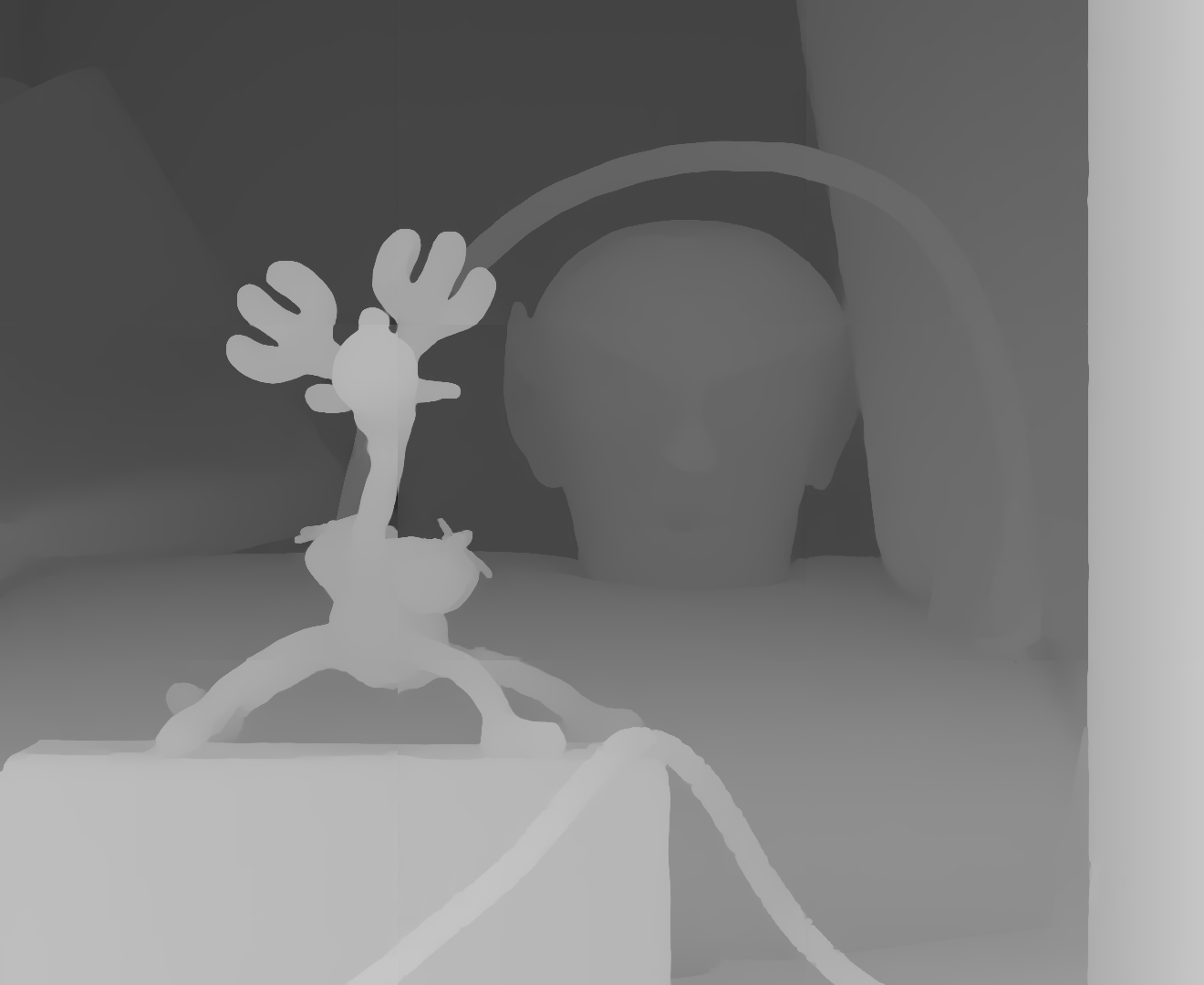}&\hspace{-13pt}
				\includegraphics[width=2.5cm]{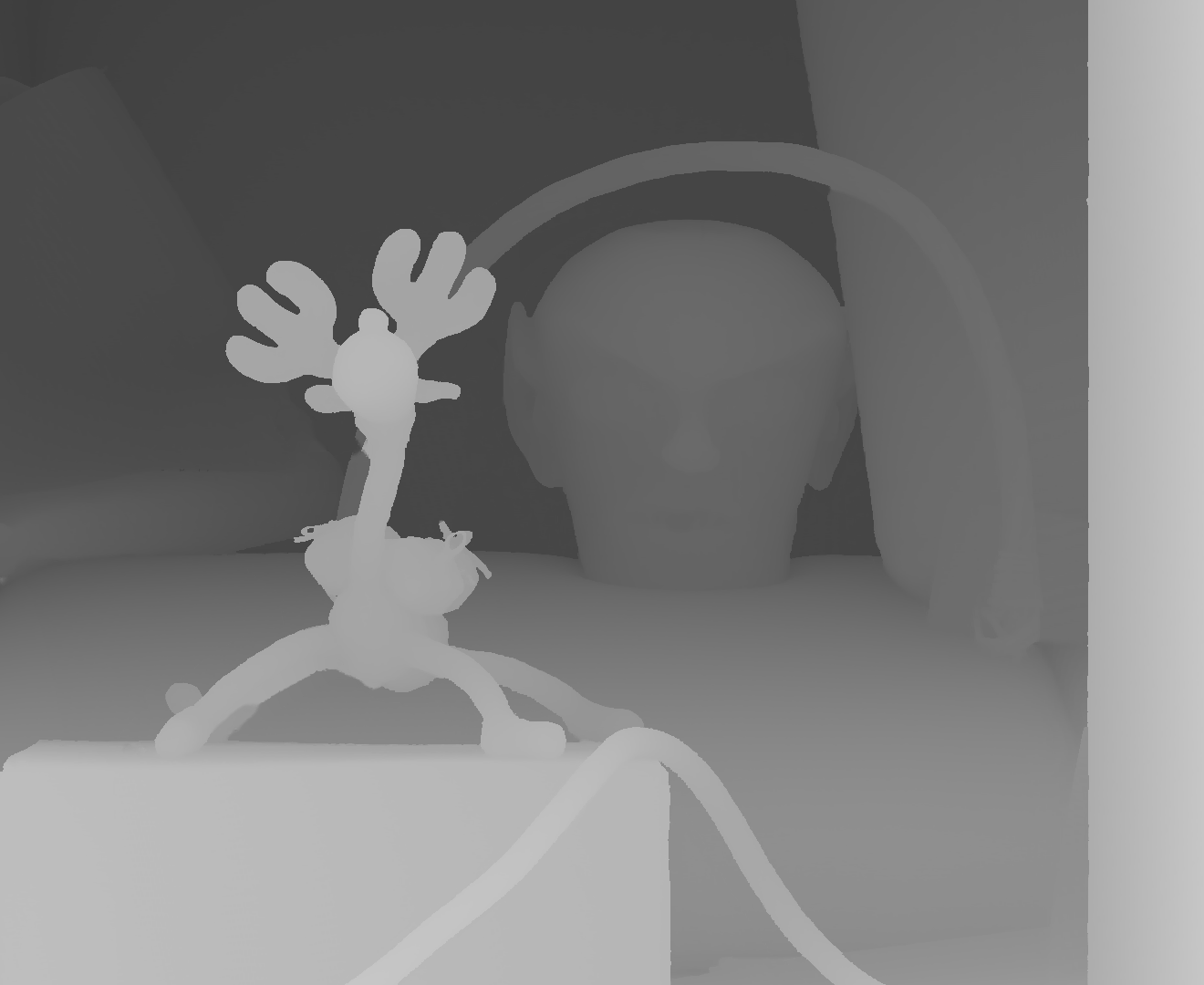}\\			
				(a) & (b) & (c) & (d) & (e) & (f) & (g)\\		
			\end{tabular}}	
			\caption{Qualitative comparison of depth map super-resolution results on noise-free \textit{Art} and \textit{Reindeer} datasets (upsampling factor = $16$). (a) RGB image, upsampled depth maps by (b) Bicubic, (c) TGV \cite{Ferstl_iccv2013}, (d) RCG \cite{Liu_tip2017}, (e) DepthSR-Net \cite{Guo_tip2019}, (f) PAG-Net. (g) Ground truth.}
			\label{fig:vis2}
		\end{figure*}

\textbf{Loss Function} Since $L_1$ loss functions works better for making the images as sharp as possible while $L_2$ is very effective for removing the noise present in the image. In this work, we trained the proposed network represented as $\mathbf{F}$ on the combination of $L_1$ and $L_2$ loss. Hence the combined loss function is 
\begin{equation}
\begin{split}
	L(\mathbf{\theta}) = \frac{1}{N}\sum\limits_{i=1}^{N}(\parallel \mathbf{F}(\mathbf{D}^l_i,\mathbf{I}^{h}_i,\mathbf{\theta})- \mathbf{D}^h_i \parallel_2 \\+ \parallel \mathbf{F}(\mathbf{D}^l_i,\mathbf{I}^{h}_i,\theta)- \mathbf{D}^h_i \parallel_1)
	\end{split}
\end{equation}                               
where $\mathbf{D}^l$ is the low-resolution depth map, $\mathbf{D}^h$ represents high-resolution depth map, $\mathbf{I}^h$ high-resolution intensity image corresponding to $\mathbf{D}^h$, $\mathbf{\theta}$ network parameters, and $N$ total number of images used for training. 

\textbf{Implementation details:}
We implemented the proposed PAG-Net using PyTorch on a PC with an i$7$-$6700$ CPU, $64$GB RAM, and a GTX $1080$Ti GPU. The weights in each filter initialized from a zero-mean Gaussian distribution and biases are initialized with constant zero values. In this work, we train specific models for each upscaling factor such as $2$x, $4$x, $8$x, $16$x, respectively.
All the models trained for $30$ epochs by setting batch size as $8$, and the learning rate $1e-4$. We optimize network parameters using ADAM with $\beta_1=0.9$ and $\beta_2=0.99$. During the testing phase, we obtain the high-resolution depth map by feeding low-resolution depth map and its corresponding high-resolution color image.

\section{Experimental Results}
\label{sec:exp}

\textbf{Training details:} In order to evaluate the proposed framwork, we followed the similar way in \cite{Guo_tip2019,hui_eccv16} for training and validation data selection. We used a total number of $82$ RGB-D images for training among which $58$ images are taken from MPI sintel depth dataset \cite{Daniel_eccv2012} and the remaining $34$ images are from Middlebury dataset \cite{Scharstein_ijcv2002}. 
Since the available data is small enough for training deep networks, we augmented data with standard techniques such as flipping and rotation which leads to an augmented training dataset $8$ times larger than the original dataset. In the training phase, the HR depth maps $\mathbf{D}^h$ were cropped to $256\times 256$ image patches by overlapping sampling with a stride of $64$ for all scaling factors. At last, the augmented training data provided roughly $1,26,000$ image patches. To synthesize low-resolution depth maps, we downsampled each full-resolution image patch by bicubic downsampling with the given scaling factor.

We evaluate proposed model both quantitatively and qualitatively on six test images from Middlebury dataset \cite{Scharstein_ijcv2002}. For quantitative evaluation we choose to compute root-means-squared error (RMSE) values. In this work, we provide comparisons with state-of-the-art depth map super-resolution algorithms including filtering-based methods, i.e.,  guided filtering (GF) \cite{He_pami2013}, optimization-based techniques namely, total generalized variation (TGV) \cite{Ferstl_iccv2013}, joint intensity and depth co-sparse method (JID) \cite{Kiechle_iccv2013}, robust color-guided (RCG) \cite{Liu_tip2017}, transform and spatial domain regularization (TSDR) \cite{Jiang_tip2018} and deep learning based models, i.e., multi-scale guidance network (MSG) \cite{hui_eccv16}, multi-scale frequency reconstruction (MFR) \cite{Zuo_tcsvt2018}, residual dense network (RDN) \cite{Zuo_IS2019}, hierarchical features driven residual learning (DepthSR) \cite{Guo_tip2019}.

In Table \ref{table:sr_quant}, we report RMSE values obtained from all the state-of-the-art methods for four different upsampling factors. Note that proposed method outperforms all the existing algorithms in noise-free depth super-resolution task. The visual comparison results on \textit{Art} and \textit{Reindeer} datasets for $16$x upsampling are shown in Fig. \ref{fig:vis2}. The super-resolution results obtained from bicubic interpolation, TGV \cite{Ferstl_iccv2013}, RCG \cite{Liu_tip2017}, DepthSR-Net \cite{Guo_tip2019}, and proposed model are shown in Figs. \ref{fig:vis2} (b)-(f), respectively. Observe that results from TGV suffer from over smoothness, RCG and DepthSR-Net contains artifacts around thin structures where as proposed model recovers high-quality depth maps. The same is evident in cropped regions in Fig. \ref{fig:vis2}.

\begin{figure}[!htb]
		\centering
		\scalebox{1.0}{
			\begin{tabular}{c c c c}
				\includegraphics[width=2.0cm]{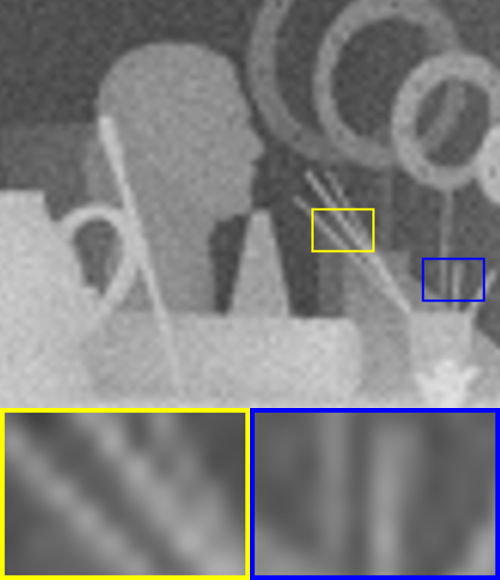}&\hspace{-13pt}
				\includegraphics[width=2.0cm]{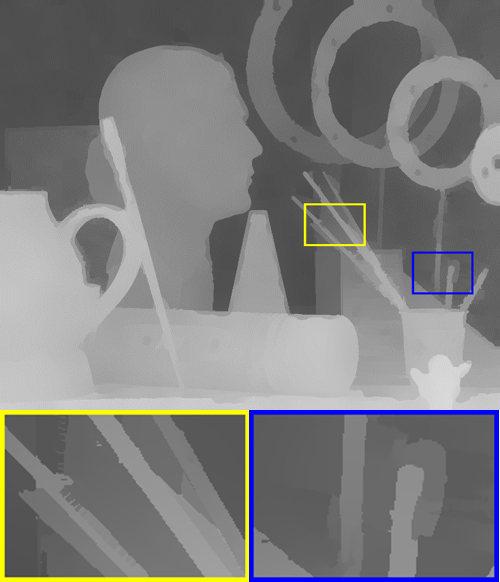}&\hspace{-13pt}
				\includegraphics[width=2.0cm]{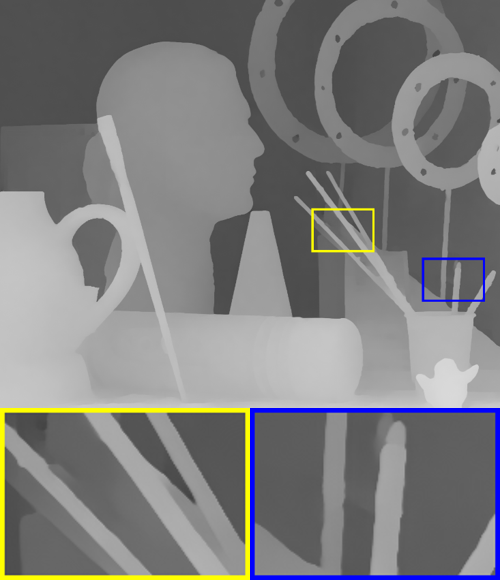}&\hspace{-13pt}
				\includegraphics[width=2.0cm]{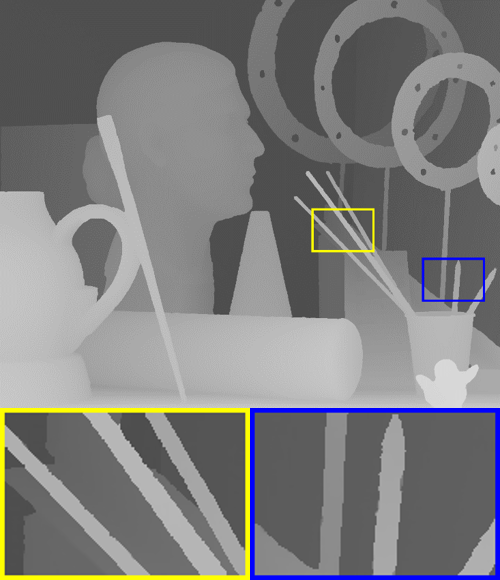}\\			
				(a) & (b) & (c) & (d)\\		
			\end{tabular}}	
			\caption{TOF-like degradation on Art dataset for the factor of $16$.}
			\label{fig:TOF}
	\end{figure}

In addition to low-spatial resolutions depth maps from TOF and Kinect sensors also contains noise in case of TOF sensors, Kinect datasets consists of missing regions. Hence, in this work we train different models for TOF-like and Kinect-like degradations. We used the same dataset as used for only super-resolution. In case of TOF, we added Gaussian noise with standard deviation $5$ similar to DepthSR-Net \cite{Guo_tip2019} and trained proposed network and named it as PAG-TOF. The visual comparison results obtained from bicubic upsampling, RCG method and proposed model for $16$x upsampling factor are shown in Fig. \ref{fig:TOF} (a)-(c), respectively.  

	

		\section{Conclusion}
		\label{sec:con}
		
		In this paper, we propose a novel method for the challenging problem of color-guided depth map super-resolution. It is based on residual dense networks and involves a novel attention mechanism to suppress the texture copying problem arises due to improper guidance by RGB images. The attention guided feature extraction module mainly involves providing the spatial attention to guidance features based on the depth features. In addition to spatial-resolution proposed model handles noise and missing regions. We provide comparisons with several state-of-the-art depth map super-resolution methods and observed that proposed model outperforms many of them. 
		

\balance


\bibliographystyle{IEEEbib}
\bibliography{icassp2020}

\end{document}